\documentclass[a4paper, 10pt, conference]{ieeeconf}      
\usepackage{FG2024}

\FGfinalcopy 
\usepackage[nolist]{acronym}
\begin{acronym}[PHOENIX14\textbf{T} ]

\acrodefplural{rnn}[RNNs]{Recurrent Neural Networks}
\acrodefplural{cnn}[CNNs]{Convolutional Neural Networks}
\acrodefplural{hmm}[HMMs]{Hidden Markov Models}
\acrodefplural{gru}[GRUs]{Gated Recurrent Units}
\acrodefplural{crf}[CRFs]{Conditional Random Fields}
\acrodefplural{gan}[GANs]{Generative Adversarial Networks}
\acrodefplural{gpu}[GPUs]{Graphic Processing Units}

\acrodefplural{mdn}[MDNs]{Mixture Density Networks}

\acro{btg}[BTG]{Bracketing Transduction Grammar}
\acro{bpe}[BPE]{Byte Pair Encoding}
\acro{bsl}[BSL]{British Sign Language}
\acro{bleu}[BLEU]{Bilingual Evaluation Understudy}
\acro{blstm}[BLSTM]{Bidirectional Long Short-Term Memory}
\acro{bslcpt}[BSLCP\textbf{T}]{BSL Corpus \textbf{T}}
\acro{cnn}[CNN]{Convolutional Neural Network}
\acro{crf}[CRF]{Conditional Random Field}
\acro{cslr}[CSLR]{Continuous Sign Language Recognition}
\acro{ctc}[CTC]{Connectionist Temporal Classification}
\acro{c4a}[C4A]{Content4All}
\acro{dl}[DL]{Deep Learning}
\acro{dgs}[DGS]{German Sign Language - Deutsche Gebärdensprache}
\acro{dsgs}[DSGS]{Swiss German Sign Language - Deutschschweizer Geb\"ardensprache}
\acro{dtw}[DTW]{Dynamic Time Warping}
\acro{dtwmje}[DTW-MJE]{Dynamic Time Warping Mean Joint Error}
\acro{fc}[FC]{Fully Connected}
\acro{ff}[FF]{Feed Forward}
\acro{fps}[fps]{frames per second}
\acro{gan}[GAN]{Generative Adversarial Network}
\acro{gpu}[GPU]{Graphics Processing Unit}
\acro{gru}[GRU]{Gated Recurrent Unit}
\acro{gtpt}[G2PT]{Gloss-to-Pose Transformer}
\acro{gtp}[G2P]{Gloss-to-Pose}
\acro{gts}[G2S]{Gloss-to-Sign}
\acro{gs}[GS]{Gloss Selection}
\acro{gr}[GR]{Gloss Reordering}
\acro{gt}[GT]{ground truth}
\acro{hmm}[HMM]{Hidden Markov Model}
\acro{hpe}[HPE]{Hand Pose Enhancer}
\acro{hoh}[HOH]{Hard of Hearing}
\acro{hns}[HamNoSys]{Hamburg Notation System}

\acro{isl}[ISL]{Irish Sign Language}
\acro{lstm}[LSTM]{Long Short-Term Memory}
\acro{mha}[MHA]{Multi-Headed Attention}
\acro{mtc}[MTC]{Monocular Total Capture}
\acro{mse}[MSE]{Mean Squared Error}
\acro{mdn}[MDN]{Mixture Density Network}
\acro{mdgs}[mDGS]{Meine DGS Annotated}
\acro{mdgst}[mDGS\textbf{T}]{meineDGS\textbf{T}}
\acro{mdgsth}[mDGS\textbf{T}-\textbf{H}]{meineDGS\textbf{T}-\textbf{HARD}}
\acro{mdgste}[mDGS\textbf{T}-\textbf{E}]{meineDGS\textbf{T}-\textbf{EASY}}
\acro{mt}[MT]{Machine Translation}

\acro{nmt}[NMT]{Neural Machine Translation}
\acro{nlp}[NLP]{Natural Language Processing}
\acro{nar}[NAR]{Non-AutoRegressive}
\acro{nsvq}[NSVQ]{Noise Substitution Vector Quantization}
\acro{ph12}[PHOENIX12]{RWTH-PHOENIX-Weather-2012}
\acro{ph14}[PHOENIX14]{RWTH-PHOENIX-Weather-2014}
\acro{ph14t}[PHOENIX14\textbf{T}]{RWTH-PHOENIX-Weather-2014\textbf{T}}
\acro{pof}[POF]{Part Orientation Field}
\acro{pos}[POS]{Part Of Speech}
\acro{pt}[PT]{Progressive Transformer}
\acro{paf}[PAF]{Part Affinity Field}
\acro{pttt}[P2TT]{Pose to Text Transformer}
\acro{ptt}[P2T]{Pose to Text}
\acro{pts}[P2S]{Pose to Sign}
\acro{ptgtt}[P2G2T]{Pose to Gloss to Text}
\acro{pca}[PCA]{Principal Component Analysis}
\acro{relu}[RELU]{Rectified Linear Units}
\acro{rnn}[RNN]{Recurrent Neural Network}
\acro{rouge}[ROUGE]{Recall-Oriented Understudy for Gisting Evaluation}
\acro{vqvae}[VQ-VAE]{Vector Quantized Variational Autoencoder}
\acro{vq}[VQ]{Vector Quantisation}
\acro{vae}[VAE]{Variational Autoencoder}
\acro{sgd}[SGD]{Stochastic Gradient Descent}
\acro{sla}[SLA]{Sign Language Assessment}
\acro{slr}[SLR]{Sign Language Recognition}
\acro{slt}[SLT]{Sign Language Translation}
\acro{slp}[SLP]{Sign Language Production}
\acro{smt}[SMT]{Statistical Machine Translation}
\acro{slo}[SLO]{Spoken Language Order}
\acro{so}[SO]{Sign Language Order}
\acro{snr}[S\&R]{Select and Reorder}
\acro{sse}[SSE]{Sign Supported English}
\acro{stt}[S2T]{Sign-to-Text}
\acro{stgtt}[S2G2T]{Sign-to-Gloss-to-Text}
\acro{ttgt}[T2GT]{Text-to-Gloss Transformer}
\acro{ttpt}[T2PT]{Text-to-Pose Transformer}
\acro{ttp}[T2P]{Text-to-Pose}
\acro{ttg}[T2G]{Text-to-Gloss}
\acro{tth}[T2H]{Text-to-HamNoSys}
\acro{ttgth}[T2G2H]{Text-to-Gloss-to-HamNoSys}
\acro{ttgtp}[T2G2P]{Text-to-Gloss-to-Pose}
\acro{tts}[T2S]{Text to Speech}
\acro{ttsse}[T2SSE]{Text to Sign Supported English}
\acro{wer}[WER]{Word Error Rate}
\acro{wmt14}[WMT2014]{WMT2014 German-English}

\end{acronym}

\usepackage{graphicx}
\usepackage{amsmath}
\usepackage{multirow}
\usepackage{booktabs}
\usepackage{tabularray}
\usepackage{longtable}
\usepackage{amssymb}
\usepackage{subfigure}
\usepackage[export]{adjustbox}
\usepackage{hyperref}
\usepackage[capitalise]{cleveref}
\usepackage{float}

\IEEEoverridecommandlockouts                              
\def\FGPaperID{195} 

\title{\LARGE \bf
A Data-Driven Representation for Sign Language Production
}

\author{\parbox{16cm}{\centering
    {\large Harry Walsh$^1$, Abolfazl Ravanshad$^2$, Mariam Rahmani$^2$ and Richard Bowden$^1$}\\
    {\normalsize
    $^1$ CVSSP, University of Surrey, Guildford, United Kingdom\\
    $^2$ OmniBridge.ai, an Intel Venture, USA}}
    \thanks{This work was supported by Intel, the SNSF project ‘SMILE II’ (CRSII5 193686), the European Union’s Horizon2020 programme (‘EASIER’ grant agreement 101016982) and the Innosuisse IICT Flagship (PFFS-21-47). This work reflects only the author's view and the Commission is not responsible for any use that may be made of the information it contains. }
}

\usepackage{fancyhdr}
\begin{document}

\ifFGfinal
\thispagestyle{empty}
\pagestyle{empty}
\else
\author{Anonymous FG2024 submission\\ Paper ID \FGPaperID \\}
\pagestyle{plain}
\fi
\maketitle
\thispagestyle{fancy} 

\begin{figure*}[htbp]
    \centering
    \caption{A overview of our approach to \acf{slp}. Showing 1) the source spoken language sentence, 2) our intermediate representation of sign, 3) the synthesized sequence of signing, and, 4) the original video.}
    \includegraphics[width=0.85\textwidth]{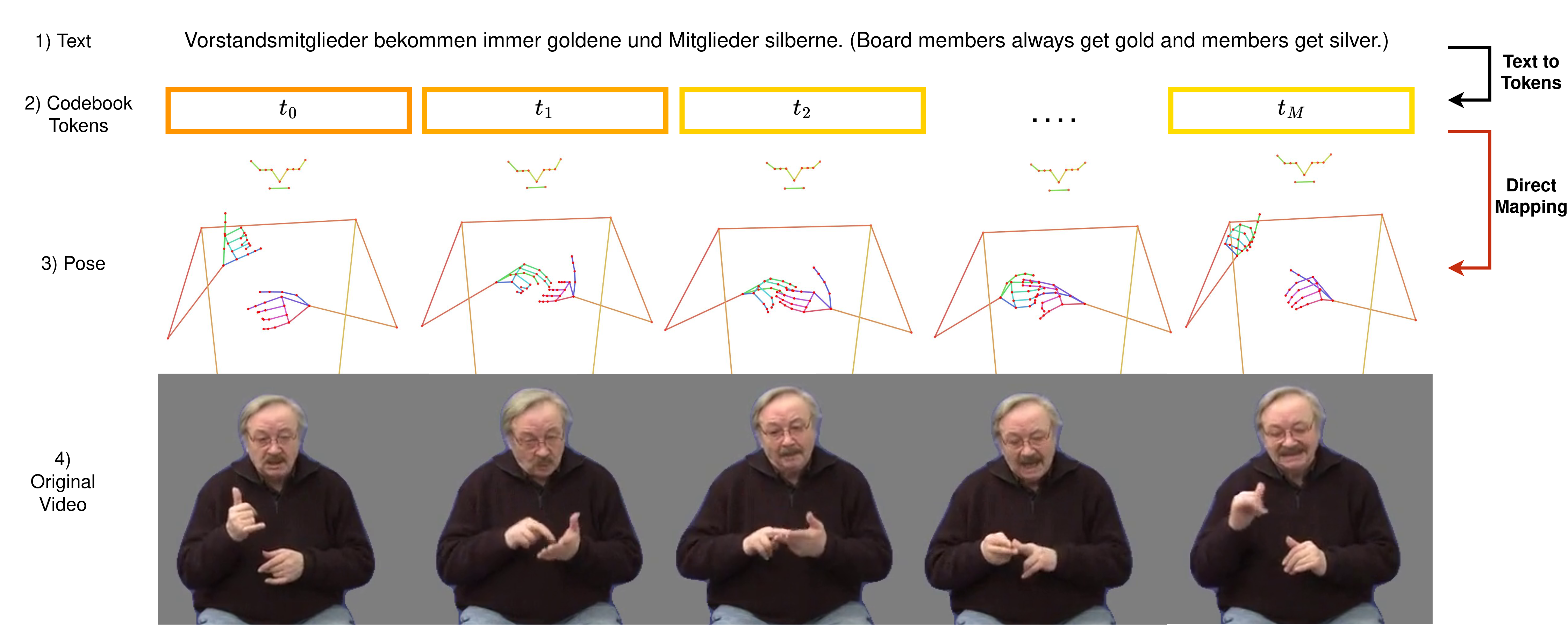}
    \label{fig:overveiw}
\end{figure*}

\begin{abstract}
Phonetic representations are used when recording spoken languages, but no equivalent exists for recording signed languages. As a result, linguists have proposed several annotation systems that operate on the gloss or sub-unit level; however, these resources are notably irregular and scarce.

Sign Language Production (SLP) aims to automatically translate spoken language sentences into continuous sequences of sign language. However, current state-of-the-art approaches rely on scarce linguistic resources to work. This has limited progress in the field. This paper introduces an innovative solution by transforming the continuous pose generation problem into a discrete sequence generation problem. Thus, overcoming the need for costly annotation. Although, if available, we leverage the additional information to enhance our approach. 

By applying Vector Quantisation (VQ) to sign language data, we first learn a codebook of short motions that can be combined to create a natural sequence of sign. Where each token in the codebook can be thought of as the lexicon of our representation. Then using a transformer we perform a translation from spoken language text to a sequence of codebook tokens.  Each token can be directly mapped to a sequence of poses allowing the translation to be performed by a single network. Furthermore, we present a sign stitching method to effectively join tokens together. We evaluate on the RWTH-PHOENIX-Weather-2014\textbf{T} (PHOENIX14T) and the more challenging meineDGS\textbf{T} (mDGS) datasets. An extensive evaluation shows our approach outperforms previous methods, increasing the BLEU-1 back translation score by up to 72\%.  

\end{abstract}

\section{INTRODUCTION}

\label{sec:intro}
Sign language is a rich and complex form of communication that relies on visual-spatial elements rather than spoken words \cite{stokoe1980sign}. It serves as the primary mode of communication for the deaf community \cite{sutton1999linguistics}.

Sign languages are composed of cheremes, analogous to the phonemes found in spoken languages \cite{laver2001linguistic}. Cheremes from the Greek word for hand, describe features such as handshape, orientation, location, movement and non-manual expressions. These fundamental units can be combined to create a natural sequence of signing. When transcribing sign language, linguists commonly employ sub-unit and gloss\footnote{Gloss is the written word associated with a sign} level representations \cite{hanke2004hamnosys, sutton2022lessons}. Unfortunately, curating transcriptions is time-consuming and costly, and as a result, linguistic resources are often limited or even non-existent \cite{bragg2019sign}.

\acf{slp} is the task of translating from a spoken language sentence to a continuous sign language sequence. To facilitate natural communication, \ac{slp} must include both manual and non-manual components \footnote{Manual components include hand shape and motion. While non-manual components include facial expressions, body movements, eye gaze etc.}. Previous works have attempted to learn a direct mapping from \ac{ttp}. However, they suffer from regression to the mean \cite{saunders2020progressive}. Alternative two-step methods (\ac{ttgtp}) rely on expensive linguistic annotation \cite{gu2022vector, huang2021towards, saunders2020adversarial}. 

In this paper, we propose creating a data-driven representation of sign language that can be used as a replacement for expensive linguistic annotation. Our approach learns a codebook of motions from continuous 3D pose data using a \ac{nsvq} model \cite{vali2022nsvq}. The codebook can be considered the lexicon of our new representation and used to tokenise a continuous pose sequence into a sequence of discrete codes. As depicted in \cref{fig:overveiw}, we then tackle the problem as a traditional sequence-to-sequence task, translating from a spoken language sentence (\cref{fig:overveiw}.1) to a sequence of codebook tokens (\cref{fig:overveiw}.2). Unlike the previous two-step approaches our intermediary representation can be directly mapped to a sequence of poses (\cref{fig:overveiw}.3) and includes non-manual key points. Furthermore, we show how our representation can be enhanced when limited linguistic annotation is available, and by introducing a novel stitching module we create more natural continuous signing.

We show state-of-the-art results on \ac{ph14t} and the more challenging \ac{mdgs} dataset. An extensive ablation study reveals the effectiveness of our approach compared to previous works. Increasing the back translation scores by up to 72\%. Finally, we share quantitative results. 

We can summarise the contribution of this paper as: 
\begin{itemize}
    \item A novel architecture for creating a data-driven representation of sign language. 
    \item Sign stitching, a method to improve the back translation performance.
    \item A contrastive learning approach that enhances the representation.
    \item State-of-the-art \ac{slp} performance on \ac{ph14t} and the more challenging \ac{mdgs} dataset. 
\end{itemize}


\section{RELATED WORK}
\label{sec:related_word}
\subsection{\ac{slt}}
Over the past three decades, computational sign language translation has remained an active area of interest \cite{tamura1988recognition}. Initial research focused on isolated \ac{slr}, where a \ac{cnn} is used to classify a single instance of a sign \cite{Lecun_Gradient_based_lrn}. Advances in the field led to \ac{cslr}, which requires both the segmentation of a video into its constituent signs and their classification \cite{KOLLER2015108}. Since then the more challenging task of \ac{stt} was introduced \cite{camgoz2018neural}, where continuous sign language sequences are directly converted into spoken language sentences. However, it is shown that translating via a gloss intermediary gives better performance (\ac{stgtt}. Transformers \cite{vaswani2017attention} have been applied to the problem, and demonstrated state-of-the-art performance \cite{camgoz2020sign}. Thus, we utilise this architecture to evaluate the performance of our approach, similar to \cite{saunders2020adversarial, saunders2020progressive, saunders2021mixed}.   

\subsection{\acf{slp}}

\ac{slp} is the task of translating from spoken language sentences to a continuous sequence of sign language. Early approaches to \ac{slp} use animated avatars with a dictionary lookup \cite{glauert2006vanessa, karpouzis2007educational, mcdonald2016automated}. The first deep \ac{slp} pipeline broke down the task into three steps, first, a translation from \ac{ttg} followed by a second \ac{gtp} translation and finally a mapping from \ac{pts} \cite{stoll2020text2sign}. Saunders et al. introduced the Progressive transformer \cite{saunders2020progressive} an encoder-decoder transformer architecture that generates a continuous pose sequence given a spoken language sentence (\ac{ttp}), simplifying the \ac{slp} pipeline. They demonstrated that better translation results were achieved using a gloss intermediary (\ac{ttgtp}). However, the approach suffers from regression to the mean, resulting in under-articulated signing. In an attempt to reduce the problem, adversarial training and \ac{mdn} are added to the model \cite{saunders2020adversarial}. Alternatively, Huang et al. proposed a non-autoregressive \ac{gtp} architecture \cite{huang2021towards}, which produces the sign sequence in a single step.

Alternative representations to gloss have been explored namely the \ac{hns} \cite{hanke2004hamnosys} and SignWriting \cite{sutton2022lessons}. \ac{hns} is a transcription system that is used to describe sign language at the phonetic level, where each sign consists of a description of the initial posture and the action over time. \ac{hns} can be mapped directly to an avatar making it a suitable alternative for gloss in the \ac{slp} pipeline \cite{kaur2016hamnosys}. Furthermore, Walsh et al. defined the task of \ac{tth} and showed improved translation performance by using it as an intermediate representation \cite{walsh2022changing}. A similar translation task has been explored with translating text to SignWriting \cite{jiang2022machine} and animating from SignWriting \cite{bouzid2013avatar}.

Modern deep learning is heavily dependent on data, approximately 15 million sentence pairs are required before deep learning starts to outperform statistical approaches \cite{koehn2017six}. In contrast, sign language datasets are limited. For example, \ac{mdgs} has only 50k parallel sentences with gloss and \ac{hns} annotations \cite{dgscorpus_3}. These annotation systems are time-consuming and costly to create, and this has limited the size of the available datasets. Therefore, in this paper, we propose learning a representation from 3D pose data that can be used as a substitute for gloss or \ac{hns}. But unlike gloss, our representation can be mapped directly to a sequence of poses, removing the need for expensive annotation. But to build a discrete vocabulary we need to quantise the data. 

\subsection{Vector Quantized Models}

Kingma et al. \cite{kingma2013auto} introduced the first \ac{vae} and showed impressive results. However, they struggled to capture fine-grained structures. Van Den Oord \cite{oord2017neural} improved on this by introducing the \ac{vqvae} architecture. The model integrates \ac{vq} into the latent space of a \ac{vae}. Forcing the embedding space of the \ac{vae} to be discrete, allowing state-of-the-art image and audio generation. Since then, \ac{vq} has been applied to several problems including music generation using a hierarchical \ac{vqvae} \cite{dhariwal2020jukebox}, speech synthesis using self-supervised training \cite{baevski2020wav2vec}, and more recently image generation using a diffusion model \cite{gu2022vector}. The original \ac{vqvae} architecture used an argmin operation to select the closest matching codebook entry. As a result, the model used straight-through gradient estimation to make the model differentiable. Consequently, the model uses three separate losses to train: a reconstruction loss, a codebook loss and a commitment loss. Kaiser et al. \cite{kaiser2018fast} used an exponentially moving average to update the codebook. This simultaneously helped stabilise training and reduced the required loss functions to two. Vali et al. \cite{vali2022nsvq} then reduced the required losses to one by introducing the \ac{nsvq} technique. Here the vector quantization error is approximated by substituting it for a product of the original error and a normalised noise vector. The result allows for end-to-end training of the model showing faster convergence compared to straight-through estimation and exponential moving averages models. Therefore, we adopt this architecture in our approach.

\ac{vq} has been previously applied to the \ac{slp} problem \cite{hwang2023autoregressive, xie2022vector}. Xie et al. broke the human skeleton into three separate codebooks and used a diffusion model with codeUnet \cite{xie2022vector} to translate from \ac{gtp}. The approach still relies on expensive linguistic annotation, and qualitative results show a lack of detail in the hands resulting in under-articulated signing. In contrast, we propose approaching the task using a transformer to construct the codebook and perform the translation. We believe that the attention mechanism is more adept to modelling long-range dependencies and the change in order between the source and target. Consequently, we apply our approach to the more challenging task of \ac{ttp} translation and show higher back translation scores.

\section{METHODOLOGY}
\label{sec:methodology}
\begin{figure*}[!ht]
    \centering
    \caption{An overview of the architecture used in our approach. Showing a) The Codebook training architecture and b) the Text-to-Codebook Tokens Translation architecture.}
    \includegraphics[width=0.9\textwidth]{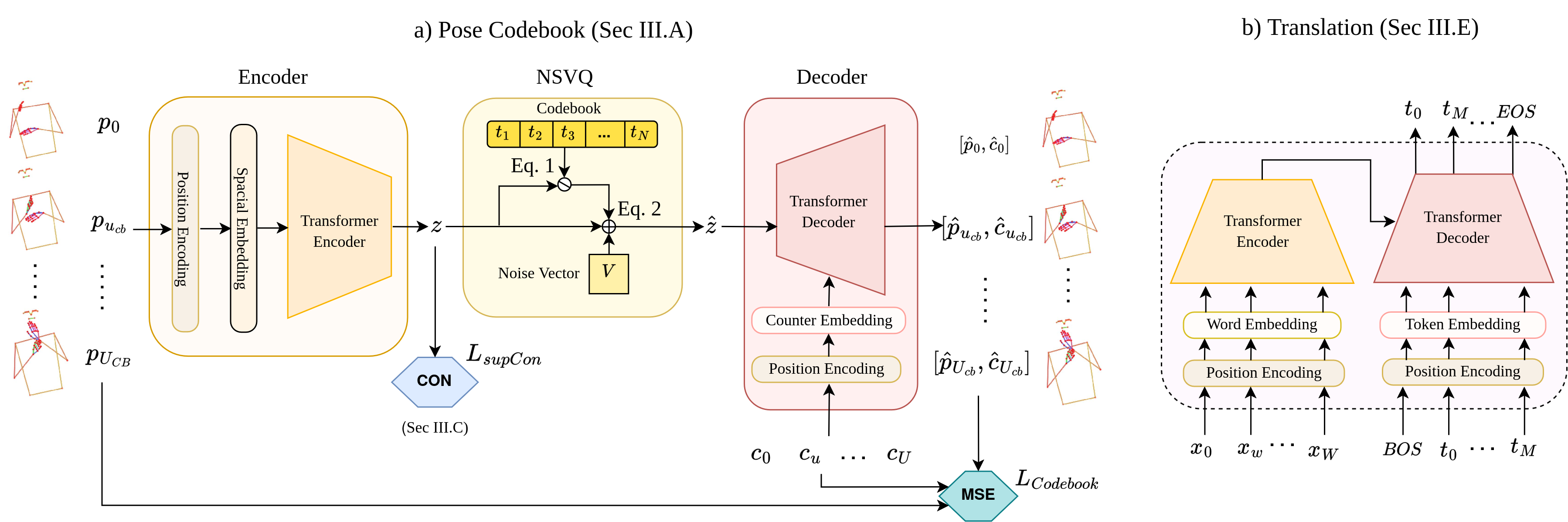}
    \label{fig:architecture}
\end{figure*}

The aim of \ac{slp} is to enable seamless translation from spoken to signed languages. To accomplish this, we convert a source spoken language sequence, denoted as \(X = (x_{1},x_{2},...,x_{W})\) with W words, into a continuous sequence of poses, denoted as \(P = (p_{1}, p_{2},...,p_{U})\) with U frames. Where each pose consists of \(J\) joints in \(D\) dimensional space e.g. $p_{i} \in \mathbb{R}^{J \times D}$. \ac{slp} is a significant challenge, considering that the target length is substantially greater than the source, such as \(U >> W\). This inherent difficulty persists even when employing state-of-the-art sequence-to-sequence models for the translation task \cite{vaswani2017attention}.
To overcome this, we first learn a codebook of tokens, that each represent a short sequence of signing and can be directly mapped back to a sequence of poses. Then we perform a translation from spoken language text to a sequence of latent codes, as shown in \cref{fig:overveiw}.1 to \ref{fig:overveiw}.2. The individual architectures of this pipeline are illustrated in \cref{fig:architecture}, and we elaborate on each stage in the subsequent sections.

\subsection{Pose Codebook}
\label{sec:pose_codebook}

The objective of the codebook is to learn a set of motions from a dataset of continuous signing. Our approach employs a transformer encoder-decoder architecture with a \ac{nsvq}. Next, we explain each module in turn, following \cref{fig:architecture}.a from left to right;  

\textbf{Encoder:} Given a short sequence of poses, \(P = (p_{1}, p_{2},...,p_{U_{cb}})\) with \(U_{cb}\) frames, we add positional encoding to each pose. We then embed the sequence using a linear layer which acts only in the spatial dimension. Then the sequence is passed to the spatial-temporal transformer encoder, allowing the network to learn long-range dependencies within the sequence. The embedded features can be defined as $z \in \mathbb{R}^{U_{cb} \times H}$, where H is the embedding size. Note we train each codebook entry to represent a sub-unit of a full continuous sequence, hence \(U_{cb} << U\).   

\textbf{\acs{vq}:} The \ac{nsvq} codebook learns a set of tokens from the encoder, we denote the codebook as \(CB = [t_{1}, t_{2},..., t_{N}]\), where N is the number of tokens in the codebook and each \(t_{i} \in \mathbb{R}^{U_{cb} \times H}\). Therefore, the length of each pose sequence, \(U_{cb}\), determines how many frames each codebook token represents. To train this module, each output from the encoder, \(z\), is mapped to a single codebook token \(t_{i}\). This is called \ac{vq} and is defined as;
\begin{equation} t_{i} \; ; \;  i = \arg \min_{t_{i}} || z - t_{i} ||^2  \label{eq:vq_min}\end{equation}

\cref{eq:vq_min} is non-differentiable. To overcome this the \ac{nsvq} simulates the quantization error by adding noise to the input vector, such that the simulated noise forms the same distribution as the original \ac{vq} error. The \ac{nsvq} is trained end-to-end and the output to the decoder can be defined as; 
\begin{equation} \hat{z} = z + || z - t_{i}|| * \dfrac{V}{|| V ||} \label{eq:nsvq} \end{equation}
Where V is a normally distributed noise source. \cref{fig:architecture}.a (NSVQ) depicts how \cref{eq:vq_min} and (\ref{eq:nsvq}) are used during training.

\textbf{Decoder:} The decoder learns to reconstruct the original pose sequence from the quantized embedding. Here we use counter decoding from Saunders et al. \cite{saunders2020progressive} to drive the decoder, and therefore, we use non-autoregressive decoding. Meaning a sequence is processed in a single step, for reduced computational cost and faster inference speeds. We find using this approach outperforms a simple multilayer perceptron on reconstruction error. The value of the counter is defined as; 
\begin{equation} c_{u} = \dfrac{u+1}{U_{cb}}\end{equation}
Where \(u\) is the current position in the sequence and \(U_{cb}\) is the total sequence length. As shown in \cref{fig:architecture}.a (Decoder), we add positional encoding and use a linear layer to embed the counter values. We then apply a spatial-temporal transformer decoder, that uses cross and self-attention to produce the output embedding. From this, we project the embedding back to the pose and counter values using two linear layers. Our architecture is trained end-to-end using the following loss function; 
\begin{equation} L_{Codebook} = \dfrac{1}{U_{cb}} \sum_{u=1}^{U_{cb}}(P_{u} - \hat{P_{u}})^2 + \alpha(c_{u} - \hat{c_{u}})^2 \end{equation}
where \(\alpha\) is a scaling factor that we determine empirically and \(\hat{P}\),  \(\hat{c}\), are the predicted pose and counter values.

\subsection{Codebook Replacement}
Codebook collapse is a significant challenge when training codebooks with \ac{vq} \cite{dieleman2018challenge}.  This is when several tokens within the codebook are no longer selected during the quantization process, resulting in dead codebook tokens. This can occur when the data distribution of the embedding space no longer matches the tokens. Strategies exist to detect and replace these dead tokens \cite{gersho2012vector, vali2022nsvq, zeghidour2021soundstream}. 

We employ two replacement strategies to reduce dead codebook entries and evenly distribute active entries. Codebook entries whose usage fall below a threshold percentage are replaced with either, 1) a randomly selected active entry, plus a small magnitude of normal noise, or, 2) a randomly selected embedding from the encoder, \(z\). By tracking the usage of each token over a given number of batches we can determine active tokens when the percentage used is greater than \(\beta\) and dead tokens when the percentage used is less than \(\gamma\). We set a schedule for training, initially using replacement more often and slowly decreasing the frequency throughout training. Once the learning rate decreases past a given factor we stop using replacement allowing the network to fine tune its parameters.    

\subsection{Contrastive Learning}
\label{sec:contrastive_learning}

When linguistic annotation is available we apply an additional contrastive loss. Specifically, we add a supervised contrastive loss \cite{khosla2020supervised} to the encoder of the codebook transformer, as shown in \cref{fig:architecture}. This makes use of gloss labels and time stamps to tag each input sequence with its corresponding gloss ID. For long input sequences that contain frames from multiple glosses, we select the most common ID. The contrastive loss pulls sequences belonging to the same gloss together, while simultaneously pushing apart sequences belonging to different glosses. We hypothesise that the additional loss allows the encoder to overcome the natural variation between signers, helping the model become person-invariant. We define the loss as,
\begin{equation}
    L_{supCon} = \sum_{i=0}^{I} \dfrac{-1}{|A(i)|} \sum_{a=0}^{A(i)}(\dfrac{exp(z_{i} \cdot z_{a} / \tau)}{\sum_{b=1}^{B(i)} exp(z_{i} \cdot z_{b} / \tau)})
\end{equation}
Here \(i\) is the index of the anchor. \(A(i)\) is the set of indices that correspond to positive samples in the batch and \(|A(i)|\) is the number of samples in a batch. While, \(B(i)\), is the set of the negative samples. \(\tau\) is a scalar temperature parameter. We define a sequence to be positive if it shares the same gloss ID with the anchor, while we define a negative sample if it has a different ID. The contrastive loss is scaled by \(\delta\) before being added to the codebook loss. Therefore, the total loss is defined as;
\begin{equation}
    L_{Total} = L_{Codebook} + \delta L_{supCon}
\end{equation}

\subsection{Pose Sequence Tokenization}

To perform a translation from text to tokens, we first tokenize the continuous pose sequence. We build the codebook to be a sub-unit representation. Thus, given a continuous sequence of poses, \(P\), we create a sequence of tokens, \(T = (t_{1}, t_{2},...,t_{M})\) where M is the number tokens, which corresponds to the length of the original sequence, such that \(M = \lfloor U_{cb} / U \rfloor\). Therefore, when tokenizing a sequence we lose any tailing frames. 
We freeze the encoder and codebook and pass each segment through the encoder. To find the corresponding token we then apply \cref{eq:vq_min}.

\textbf{De-tokenization:} A mapping between the codebook tokens and their corresponding pose sequences is obtained by passing each token through the decoder, such that;
\begin{equation}
P = D(T)
\end{equation}
We apply this mapping when evaluating the translation model's performance in the pose space.

\subsection{Text-to-Codebook Translation}

Given a spoken language sequence,  \(X = (x_{1},x_{2},...,x_{W})\) we aim to produce the corresponding sequence of codebook tokens, \(T = (t_{1}, t_{2},...,t_{M})\), therefore the translation model learns the conditional probability \(p(T|X)\). First, positional encoding is added to the spoken language sequence, \(X\). After it is embedded with a linear layer and passed through the encoder giving the context embedding used by the decoders cross attention layers. We apply autoregressive decoding, starting with the beginning of sentence token and we continue decoding until the end of sentence token is predicted, as illustrated in \cref{fig:architecture}. Similar to the encoder, positional encoding is added before each token is embedded using a linear layer.   

\subsection{Codebook Stitching}
\label{sec:codebook_stitching}

As discussed, the predicted sequence of tokens, \(T\), can be directly mapped to a sequence of poses, \(P\). However, discrepancies may arise between the final pose of one token and the initial pose of the next, resulting in discontinuities. To address this we employ linear interpolation to stitch codebook entries together, as a result we generate more natural continuous sequences. Furthermore, to maintain temporal consistency with the original sequence we fit a high-order spline curve \cite{dierckx1982algorithms} and re-sample. This maintains the number of poses in the sequence.

\section{EXPERIMENTAL SETUP}
\label{sec:setup}
\subsection{Implementation Details}
In our experiments, we search for the best hyper-parameters and find the following settings the most effective. We build our encoder-decoder translation model using a single layer with four heads, opting for an embedding size of 512 and a feed-forward size of 1024. The resultant architecture contains 7.8 million parameters. When decoding we employ a beam search algorithm with a size of 5 and a length penalty of 2.0. 

Our codebook model consists of a smaller encoder-decoder that has 1.2 million parameters. The model has 2 layers with 4 heads and is built with an embedding and feed-forward size of 128. We set a codebook replacement threshold of 0.1\%. Initially, we conduct replacement once per epoch and gradually reduce the frequency by 10 every 50 epochs. We set the initial learning rate to \(10^{-4}\) and stop the codebook replacement once the learning rate reaches \(10^{-6}\).

Both models employ dropout with a probability of 0.1 \cite{srivastava2014dropout}. We use Relu activation between the layers and apply pre-layer normalisation for regularisation and training stability. We train with a reduce on plateau scheduler with a patience of 5 and a decrease factor of 0.9. To initialize the transformer encoder and decoder layers we employ a Xavier initializer \cite{glorot2010understanding} with zero bias and Adam optimization \cite{kingma2014adam}. The learning rate is initially set to \(10^{-4}\) and we train the model till convergence. Our translation model code base comes from the Kreutzer et al. NMT toolkit, JoeyNMT \cite{kreutzer2019joey} and is implemented using Pytorch \cite{NEURIPS2019_9015}. 

For comparison on the \ac{mdgs} dataset, we train two variants of the progressive transformer till convergence with the settings presented in \cite{saunders2020progressive}.

\subsection{Dataset}
To assess our models, we employ the Public Corpus of German Sign Language, 3rd release, the \ac{mdgs} dataset \cite{dgscorpus_3}, and the \ac{ph14t} dataset (as introduced by Camgoz et al., 2018 \cite{camgoz2018neural}). The \ac{mdgs} dataset is comprised of aligned spoken German sentences and gloss sequences, obtained from unconstrained dialogues between two native deaf signers. Whereas the \ac{ph14t} dataset comes from German weather broadcasts and includes 8257 sequences performed by 9 signers. Resulting in 1066 signed glosses and a spoken language vocabulary of 2887. Notably, the \ac{mdgs} is 7.5 times larger than \ac{ph14t}, featuring 330 deaf participants engaging in free-form signing and a spoken language vocabulary of 18,457. We follow the formatting conventions set out by Saunders et al. in \cite{saunders2021signing}. In addition, we eliminate gloss variant numbers to mitigate singletons when translating \ac{gtp}, in \cref{sec:state_of_the_art_comp}.

To obtain the pose from the original video, we employ Mediapipe to extract 61 2D keypoints (comprised of 21 for each hand, 9 for the body, and 10 for the face) \cite{lugaresi2019mediapipe}. To ensure the accurate elevation from 2D to 3D, we adopt the methodology outlined in \cite{10193629}. This approach utilises forward kinematics and a neural network to predict both bone lengths and angles from the 2D pose. Each pose is represented as a hierarchical tree, enforcing physical limits to constrain the pose and ensure it remains valid. When two camera angles are available (such as in the \ac{mdgs} dataset) we extract 3D pose using the method above and run an additional optimization, minimizing the error between the two predicted poses. The \ac{mdgs} and \ac{ph14t} datasets are captured at 50 and 25 \ac{fps}, respectively. We reduce the frame rate by subsampling each pose sequence by a factor of 3, this removes redundant information and speeds up training.   

\subsection{Evaluation Metrics}
For evaluation purposes, we employ the back translation metric \cite{saunders2020progressive}. For which we use the state-of-the-art \ac{cslr} architecture (Sign Language Transformers \cite{camgoz2020sign}), the same as \cite{huang2021towards, saunders2020adversarial, saunders2021mixed, xie2022vector}. The model is a transformer encoder-decoder which predicts a spoken language sentence given a pose sequence. The model employs \ac{ctc} loss \cite{graves2006connectionist} as additional supervision to predict gloss tokens. We train a model for each dataset and freeze the parameters so results are consistent across runs. We compute BLEU scores (BLEU-1,2,3, and 4) \cite{papineni2002bleu} and ROUGE scores \cite{lin2004rouge} against the original input text or gloss.  

To evaluate the accuracy of the poses we use \ac{dtwmje}, this metric aligns two-time series by stretching or compressing them locally in time to find the optimal match, minimizing the overall distance between the \ac{gt} and predicted. Thus, we first calculate the index alignment;
\begin{equation} A_{i,j} = DTW(p_u, \hat{p}_u) \end{equation}
After the alignment, we compute the mean joint error between the two;
\begin{equation} \text{DTW-MJE}= \sum_{u=0}^{U}|p_u[A_{i}]-\hat{p}_u[A_{j}]| \label{eq:dtwmje}\end{equation}
Note we normalise the skeletons between the range of zero and one before calculating \ac{dtwmje}.

\begin{table*}
\centering
\caption{\label{tab:ph_vocab_size}The results of translating from spoken language Text-to-Pose with different codebook vocabulary sizes on the RWTH-PHOENIX-Weather-2014\textbf{T} dataset.}
\resizebox{\linewidth}{!}{%

\begin{tabular}{r|cccccc|cccccc} 
\toprule
\multicolumn{1}{l}{PHOENIX14\textbf{T}}      & \multicolumn{6}{c}{TEST SET}               & \multicolumn{6}{c}{DEV SET}          \\
\multicolumn{1}{c|}{Vocabulary Size:} & DTW-MJE & BLEU-1 & BLEU-2 & BLEU-3 & BLEU-4 & ROUGE & DTW-MJE & BLEU-1 & BLEU-2 & BLEU-3 & BLEU-4 & ROUGE  \\ 
\midrule
500              & \textbf{0.08563}      & 24.24  & 13.11  & 9.31   & 7.41   & 25.064 & \textbf{0.08548}       & 23.71  & 12.68  & 8.79   & 6.84   & 24.19 \\
1,000             & 0.09401      & 24.78  & 13.96  & 9.85   & 7.71   & 25.90  & 0.09336       & 24.37  & 13.31  & 9.08   & 6.97   & 25.34 \\
3,000             & 0.1029       & 25.70  & 14.48  & 10.179 & 7.90  & 26.68  & 0.1039        & 25.08  & 13.72  & 9.53   & 7.316  & 26.27 \\
4,000             & 0.1047       & \textbf{27.74}  & \textbf{16.36}  & \textbf{11.75}  & \textbf{9.20}   & \textbf{27.93}  & 0.1036         & \textbf{27.85}  & \textbf{16.71}  & \textbf{12.19 } & \textbf{9.64}   & \textbf{28.87} \\
5,000             & 0.1036       & 24.71  & 13.39  & 9.39   & 7.22   & 26.03  & 0.1029        & 24.99  & 13.47  & 9.20   & 6.955  & 25.74 \\
6,000             & 0.1044       & 22.81  & 12.35  & 8.93   & 7.16   & 24.33  & 0.1045        & 25.37  & 14.65  & 10.66  & 8.36   & 27.14 \\

\bottomrule
\end{tabular}
}
\end{table*}
\begin{table*}
\centering
\caption{\label{tab:mdgs_vocab_size}The results of translating from spoken language Text-to-Pose with different codebook vocabulary sizes on the Meine DGS Annotated (mDGS) dataset.}
\resizebox{\linewidth}{!}{%

\begin{tabular}{r|cccccc|cccccc} 
\toprule
\multicolumn{1}{l}{mDGS}      & \multicolumn{6}{c}{TEST SET}               & \multicolumn{6}{c}{DEV SET}          \\
\multicolumn{1}{c|}{Vocabulary Size:} & DTW-MJE & BLEU-1 & BLEU-2 & BLEU-3 & BLEU-4 & ROUGE & DTW-MJE & BLEU-1 & BLEU-2 & BLEU-3 & BLEU-4 & ROUGE  \\ 
\midrule
500              & 0.1090          & 14.94          & 3.05          & 0.57          & 0.15          & \textbf{21.54} & 0.1078          & 14.80          & 3.01          & 0.632         & 0.21           & \textbf{21.14} \\
1,000            & \textbf{0.1071} & 14.72          & 3.04          & 0.73          & 0.27          & 21.25          & \textbf{0.1069} & 14.84          & 2.83          & 0.59          & 0.18           & 21.01          \\
2,500            & 0.1153          & \textbf{15.83} & 2.99          & 0.77          & 0.19          & 21.21          & 0.1157          & \textbf{15.71} & 2.80          & 0.72          & 0.00           & 20.82          \\
3,000            & 0.1251          & 15.08          & 2.84          & 0.70          & 0.21          & 20.66          & 0.1245          & 15.25          & 2.78          & 0.71          & 0.23           & 20.54          \\
3,500            & 0.1241          & 15.45          & \textbf{3.06} & \textbf{0.79} & \textbf{0.29} & 21.41          & 0.1227          & 15.43          & \textbf{3.01} & 0.72          & 0.21           & 20.91          \\
5,000            & 0.1288          & 14.95          & 2.63          & 0.68          & 0.24          & 21.12          & 0.1306          & 14.87          & 2.74          & \textbf{0.86} & 0.38           & 20.99          \\
10,000           & 0.1262          & 7.55           & 1.37          & 0.36          & 0.00          & 20.71          & 0.1271          & 7.58           & 1.29          & 0.31          & \textbf{0.098} & 20.58          \\
25,000           & 0.1203          & 10.23          & 1.75          & 0.34          & 0.00          & 21.49          & 0.1193          & 10.19          & 1.72          & 0.41          & 0.14           & 21.13          \\

\bottomrule
\end{tabular}
}
\end{table*}

\section{EXPERIMENTS}
\subsection{Quantitative Evaluation}
\label{sec:experiments}
In this section, we provide a quantitative evaluation of our \ac{slp} approach. Initially, we search for the optimal vocabulary size and window size for our codebooks, showing the result is dataset dependent.  Following this, we conduct an ablation study on the \ac{mdgs} dataset, demonstrating the advantages of both the contrastive loss and the stitching model. We then apply an enhanced codebook trained on the \ac{mdgs} to the \ac{ph14t} dataset. Finally, to facilitate a meaningful comparison on the \ac{ph14t}, we train a \ac{gtp} model and assess its performance against prior works.

\subsubsection{Codebook Vocabulary Size}
\label{sec:vocabulary_size}

Our first experiment searches for the best vocabulary size for each dataset. We fixed the window size to 4 frames and trained each codebook till convergence. After we freeze the codebook and use it to tokenize each dataset for translation. 

As shown in \Cref{tab:ph_vocab_size} the best vocabulary size is found to be 4,000 on \ac{ph14t}, achieving an impressive 27.85 BLEU-1 and 28.87 ROUGE score. The optimum is roughly 4 times larger than the original gloss vocabulary, suggesting the model is distinguishing between lexical variations. Whereas, on the larger dataset, \ac{mdgs}, a smaller vocabulary is found to be optimal at 2,500 as shown in \Cref{tab:mdgs_vocab_size}. Suggesting the model is finding a subunit representation given the original gloss vocabulary was approximately 4,500. At the optimal vocabulary, a reasonable score of 15.83 BlEU-1 and 21.21 ROUGE were achieved. However, the model showed limited performance on BLEU-2 to 4 metrics. This is due to the limits of the back translation model, as on the ground truth data, the model achieved only 0.8 BLEU-4 (shown in row 1 \Cref{tab:mdgs_abalation}, GT). Relative to this theoretical maximum our model performs well.

The smaller the codebook size the more data points map to a single token, as a result, tokens can suffer from regression to the mean, resulting in under-articulated signing. Hence, on both datasets, the lowest \ac{dtwmje} is achieved at small codebook sizes as each token is more likely to contain a mean pose. Therefore we choose to follow BLEU and ROUGE scores when deciding the best vocabulary size. 

On \ac{ph14t} we find our best codebook used 3985 tokens to tokenize the training data, a 99.6\% vocabulary usage. \ac{mdgs} has a similar result, with 99.8\% of tokens being used. The aggressive codebook replacement strategy effectively removes dead tokens, enabling the high codebook utility. 

\subsubsection{Codebook Window Size}
\label{sec:window_size}
\begin{table*}
\centering
\caption{\label{tab:ph_window_size}The results of translating from spoken language Text-to-Pose with different codebook window sizes on the RWTH-PHOENIX-Weather-2014\textbf{T} dataset.}
\resizebox{\linewidth}{!}{%

\begin{tabular}{r|cccccc|cccccc} 
\toprule
\multicolumn{1}{l}{PHOENIX14\textbf{T}}      & \multicolumn{6}{c}{TEST SET}               & \multicolumn{6}{c}{DEV SET}          \\
\multicolumn{1}{c|}{Window Size:} & DTW-MJE & BLEU-1 & BLEU-2 & BLEU-3 & BLEU-4 & ROUGE & DTW-MJE & BLEU-1 & BLEU-2 & BLEU-3 & BLEU-4 & ROUGE  \\ 
\midrule
2            & 0.0975          & 23.85          & 12.26          & 8.26           & 6.38          & 24.15          & 0.1004          & 23.03          & 11.85          & 7.88           & 5.89          & 24.08          \\
4            & 0.1047          & \textbf{27.74} & \textbf{16.36} & \textbf{11.75} & \textbf{9.20} & \textbf{27.93} & 0.1036          & \textbf{27.85} & \textbf{16.71} & \textbf{12.19} & \textbf{9.64} & \textbf{28.87} \\
8            & 0.0983          & 23.37          & 12.67          & 8.97           & 7.05          & 24.77          & 0.9953          & 23.32          & 12.91          & 9.34           & 7.58          & 25.34          \\
12           & 0.1000          & 21.12          & 11.27          & 8.16           & 6.58          & 23.49          & 0.0997          & 21.48          & 11.40          & 8.02           & 6.30          & 23.74          \\
16           & 0.0924          & 19.17          & 9.59           & 6.63           & 5.21          & 21.37          & 0.0917          & 19.43          & 9.83           & 6.85           & 5.36          & 22.17          \\
24           & \textbf{0.0893} & 19.07          & 9.77           & 7.00           & 5.63          & 22.33          & \textbf{0.0899} & 17.94          & 9.20           & 6.42           & 5.07          & 21.67          \\
32           & 0.0914          & 19.03          & 9.72           & 7.32           & 5.89          & 21.74          & 0.8925          & 17.82          & 8.80           & 5.83           & 4.48          & 20.93          \\

\bottomrule
\end{tabular}
}
\end{table*}
\begin{table*}
\centering
\caption{\label{tab:mdgs_window_size}The results of translating from spoken language Text-to-Pose with different codebook window sizes on the Meine DGS Annotated (mDGS) dataset.}
\resizebox{\linewidth}{!}{%

\begin{tabular}{r|cccccc|cccccc} 
\toprule
\multicolumn{1}{l}{mDGS}      & \multicolumn{6}{c}{TEST SET}               & \multicolumn{6}{c}{DEV SET}          \\
\multicolumn{1}{c|}{Window Size:} & DTW-MJE & BLEU-1 & BLEU-2 & BLEU-3 & BLEU-4 & ROUGE & DTW-MJE & BLEU-1 & BLEU-2 & BLEU-3 & BLEU-4 & ROUGE  \\ 
\midrule
2            & 0.1321          & 14.59          & 2.49          & 0.54          & 0.00          & 21.05          & 0.1326          & 14.38          & 2.37          & 0.51          & 0.14          & 20.70          \\
4            & 0.1277          & 15.45          & \textbf{3.12} & 0.70          & 0.00          & \textbf{21.38} & 0.1269          & 15.46          & \textbf{2.90} & 0.62          & \textbf{0.25} & \textbf{20.95} \\
8            & 0.1153          & \textbf{15.83} & 2.99          & \textbf{0.77} & \textbf{0.29} & 21.21          & 0.1157          & \textbf{15.71} & 2.80          & \textbf{0.71} & 0.00          & 20.82          \\
12           & 0.1147          & 14.75          & 2.72          & 0.57          & 0.21          & 20.82          & 0.1142          & 15.05          & 2.79          & 0.63          & 0.00          & 20.93          \\
16           & 0.1120          & 14.22          & 2.72          & 0.74          & 0.26          & 19.79          & 0.1115          & 14.31          & 2.54          & 0.63          & 0.23          & 19.73          \\
24           & \textbf{0.1113} & 12.05          & 1.89          & 0.44          & 0.13          & 17.95          & \textbf{0.1092} & 12.01          & 1.76          & 0.33          & 0.00          & 18.05          \\
32           & 0.1124          & 11.35          & 1.20          & 0.32          & 0.10          & 16.51          & 0.1118          & 11.34          & 1.80          & 0.35          & 0.00          & 16.62          \\

\bottomrule
\end{tabular}
}
\end{table*}

Next, we investigate the best window size for each codebook entry. The vocabulary is fixed to the optimum found in the previous experiment, 4,000 on \ac{ph14t} and 2,500 on \ac{mdgs}. On \ac{ph14t} we find a BLEU and ROUGE of 27.85 and 28.87 respectively, as shown in \Cref{tab:ph_window_size}. On \ac{mdgs} a window size of 8 frames was found to be the optimum (\Cref{tab:mdgs_window_size}). However, both correspond to 0.48 seconds of signing, since \ac{ph14t} and \ac{mdgs} were captured at 25 and 50 \ac{fps}, respectively.

Examination of both \Cref{tab:mdgs_window_size} and \Cref{tab:ph_window_size} reveals that increasing the window size beyond 4 and 8, respectively, decreased the BLEU and ROUGE scores while improving the \ac{dtwmje}. The minimum \ac{dtwmje} was observed at a window size of 24. Possibly due to the reduced number of tokens in the target sequence that led to fewer discontinuities in the pose sequence.  

\subsubsection{Ablation Study}
\begin{table*}
\centering
\caption{\label{tab:mdgs_abalation}The results of translating from Text-to-Pose with different approaches on the Meine DGS Annotated (mDGS) dataset.}
\resizebox{\linewidth}{!}{%

\begin{tabular}{r|cccccc|cccccc} 
\toprule
\multicolumn{1}{l}{mDGS}      & \multicolumn{6}{c}{TEST SET}               & \multicolumn{6}{c}{DEV SET}          \\
\multicolumn{1}{c|}{Approach:} & DTW-MJE & BLEU-1 & BLEU-2 & BLEU-3 & BLEU-4 & ROUGE & DTW-MJE & BLEU-1 & BLEU-2 & BLEU-3 & BLEU-4 & ROUGE  \\ 
\midrule
GT                                & 0.000           & 20.87          & 5.60          & 1.89          & 0.80          & 23.78          & 0.000           & 20.75          & 5.43          & 1.81          & 0.76          & 23.41          \\
\midrule
Quantization                      & 0.0612          & 17.88          & 3.59          & 0.76          & 0.22          & 20.86          & 0.0611          & 17.97          & 3.35          & 0.81          & 0.25          & 20.73          \\
PT                                & 0.2291          & 6.11           & 0.94          & 0.21          & 0.05          & 8.36           & 0.2284          & 6.22           & 0.98          & 0.17          & 0.00          & 8.44           \\
PT + GN                           & 0.2245          & 7.18           & 1.48          & 0.40          & 0.01          & 8.38           & 0.2241          & 9.22           & 1.63          & 0.38          & 0.01          & 8.57           \\
Codebook                          & 0.1153          & 15.83          & 2.99          & 0.77          & 0.19          & 21.21          & 0.1157          & 15.71          & 2.80          & 0.72          & 0.00          & 20.82          \\
Codebook + Stitching                 & 0.1135          & 16.32          & 3.12          & 0.80          & 0.00          & 20.96          & 0.1137          & 16.11          & 3.02          & 0.79          & 0.19          & 20.72          \\
Codebook + Contrastive            & 0.0882          & 17.44          & 3.66          & 0.99          & 0.35          & \textbf{22.25} & 0.0879          & 17.45          & \textbf{3.78} & \textbf{1.20} & \textbf{0.46} & 21.98          \\
Codebook + Contrastive + Stitching & \textbf{0.0865} & \textbf{17.62} & \textbf{3.72} & \textbf{1.04} & \textbf{0.39} & 22.23          & \textbf{0.0866} & \textbf{17.53} & 3.76          & 1.15          & 0.43          & \textbf{22.15} \\

\bottomrule
\end{tabular}
}
\end{table*}
\label{sec:ablation_study}

We start by sharing the results of training the back translation model (GT \Cref{tab:mdgs_abalation}). The model achieves good BLEU-1 and ROUGE scores. However, the performance was limited on BLEU-2 to 4. The \ac{mdgs} dataset is challenging with a spoken language lexicon of 29,275 (10 times that of \ac{ph14t}) which limits the back translation results. As such, row 1, GT, should be considered the upper bound for all experiments on \ac{mdgs}. 

Tokenizing a pose sequence with a codebook causes two quantization errors. Firstly, tailing frames are lost if the sequence is not a multiple of the window size, and, secondly, an error in the pose is caused by selecting the closest codebook token. As shown in \Cref{tab:mdgs_abalation} the accumulation of these errors reduced the performance from the GT (row 1) to Quantized (row 2), a decrease in BLEU-1 to 4 of 2.78, 2.08, 1.00 and 0.51 on the Dev set, a relatively small decrease in performance. 

For comparison, the following two rows (PT and PT + GN) of \Cref{tab:mdgs_abalation} show the results of training a Progressive Transformer on the same pose data with the same normalisation. Adding Gaussian noise (GN) to the input at a rate of 5 increased the BLEU-1 and ROUGE scores by 3.00 and 0.13, respectively on the Dev set (shown in row 4, PT + GN). Despite this augmentation, our baseline approach was shown to outperform both versions of the progressive transformer. Showing an impressive BLEU-1 increase of 8.65 and 6.49 on the Test and Dev set.

Now we present two additional techniques to improve the performance of our baseline model. 1) A supervised contrastive learning technique, described in \Cref{sec:contrastive_learning} and, 2) a codebook stitching approach, described in \ref{sec:codebook_stitching}. 

When training the codebook we apply an additional loss to the encoder of the model, this encourages sequences from the same gloss to have a similar embedding, while simultaneously pushing them away from sequences with a different ID. We believe this helps the model to ignore the natural variation between signers, allowing it to focus on the core similarities. As shown in \Cref{tab:mdgs_abalation} "Codebook + Contrastive" (row 6) the incorporation of the loss improved the performance on all metrics. The \ac{dtwmje} improved by 23\% and 24\% on the test and dev set, respectively, while the BLEU-1 score increased by 1.61 and 1.74. Showing that linguistic annotation can be used to further improve the approach. 

The stitching module is applied to the predicted sequence to create smoother transitions between codebook tokens, as described in \Cref{sec:codebook_stitching}. \Cref{tab:mdgs_abalation} "Codebook + Stitching" shows the stitching module increased the back translation BlEU-1 score by 0.49, while also improving the mean joint error. This also had qualitative improvements reducing the number of discontinuities in the predicted sequences, and as a result, the sequence was more realistic. To evaluate this experimentally we average the velocity of the signer's skeleton and find that the original data has a standard deviation of 0.044. In contrast, the quantized sequence has a deviation of 0.056. By applying the stitching module to the output the standard deviation moves closer to the original data to 0.039. 

\subsubsection{Cross-Corpus Codebook Study}
\begin{table}
\centering
\caption{\label{tab:mdgs_CB_phix_data}The results of translating from spoken language Text-to-Pose on the RWTH-PHOENIX-Weather-2014\textbf{T} with a codebooks trained on different dataset.}
\resizebox{0.98\linewidth}{!}{%

\begin{tabular}{r|ccc|ccc} 
\toprule
\multicolumn{1}{l}{PHOENIX14\textbf{T}}      & \multicolumn{3}{c}{TEST SET}               & \multicolumn{3}{c}{DEV SET}          \\
\multicolumn{1}{c|}{Approach:} & DTW-MJE & BLEU-1 & ROUGE & DTW-MJE & BLEU-1 & ROUGE  \\ 
\midrule
PHIX Codebook  & 0.1014   & 24.68    & 23.89 & 0.0997  & 24.40   & 24.21 \\
mDGS Codebook  & 0.1108   & 22.51    & 23.38 & 0.1106  & 22.71   & 23.92 \\
mDGS Codebook+ & 0.1064   & 24.32    & 24.03 & 0.1046  & 24.03   & 23.31 \\
\bottomrule
\end{tabular}
}
\end{table}

Here we investigate if a codebook trained on a high-resource dataset can be applied to another. We take a codebook trained on the \ac{mdgs} and use it to perform translation on the \ac{ph14t} dataset. As a baseline, we train a codebook with the same hyperparameters (\Cref{tab:mdgs_CB_phix_data} row 1, PHIX Codebook). We use two codebooks, a normal model (mDGS Codebook) and an enhanced codebook with stitching plus contrastive learning (mDGS Codebook+).

The results show codebooks can be shared across datasets, although with some reduction in performance, compared to training on the original data. Between \Cref{tab:mdgs_CB_phix_data} rows 1 and 2, we see a small decrease in BLEU-1 and ROUGE of 1.69 and 0.29, on the Dev set respectively. However, applying the enhanced codebook to the \ac{ph14t} dataset recovered the majority of the lost performance. 

\subsubsection{State-of-the-art Comparison}
\label{sec:state_of_the_art_comp}
\begin{table}
\centering
\caption{\label{tab:phix_state-of-the-art}The results of translating from Gloss-to-Pose on the RWTH-PHOENIX-Weather-2014\textbf{T} dataset.}
\resizebox{0.99\linewidth}{!}{%

\begin{tabular}{r|cccccc} 
\toprule
\multicolumn{1}{l}{PHOENIX14\textbf{T}}      & \multicolumn{6}{c}{}          \\
\multicolumn{1}{c|}{Approach:} & DTW-MJE & BLEU-1 & BLEU-2 & BLEU-3 & BLEU-4 & ROUGE  \\ 
\midrule
GT                & 0.000    & 32.41  & 20.19   & 14.41   & 11.32   & 32.96 \\
\midrule
PT \cite{saunders2020progressive}               & 0.191    & 11.45  & 7.08   & 5.08   & 4.04   & -      \\
NAT-AT \cite{huang2021towards} & 0.177    & 14.26  & 9.93   & 7.11   & 5.53   & -      \\
NAT-EA \cite{huang2021towards} & 0.146    & 15.12  & 10.45  & 7.99   & 6.66   & -      \\
PoseVQ-MP  \cite{xie2022vector} & 0.146    & 15.43  & 10.69  & 8.26   & 6.98   & -      \\
PoseVQ\_Diffusion \cite{xie2022vector} & 0.116    & 16.11  & 11.37  & 9.22   & 7.50   & -      \\ 
\midrule
G2P Ours          & \textbf{0.098}    & \textbf{25.46} & \textbf{14.40}  & \textbf{10.33}  & \textbf{8.17}   & \textbf{26.898} \\
T2P Ours          & \textbf{0.105}    & \textbf{27.74} & \textbf{16.36}  & \textbf{11.75}  & \textbf{9.20}   & \textbf{27.93} \\
\bottomrule
\end{tabular}
}
\end{table}

Finally, to compare against previous works we take our best-performing parameters found in \Cref{sec:vocabulary_size} and \ref{sec:window_size} and apply them to the \ac{gtp} task. Results for comparison are provided by \cite{xie2022vector}. We find our approach outperforms all previous methods on all metrics, including the progressive transformer \cite{saunders2020progressive}, a non-autoregressive transformer \cite{huang2021towards} and a diffusion-based approach \cite{gu2022vector}. Compared to the next best model we achieve improvements of 10.5\% and 72.2\% in \ac{dtwmje} and BLEU-1. 

Surprisingly, we find the best performance when translating from text. This could be because of two reasons, 1) gloss is not a perfect representation of sign language and lacks many essential channels (mouthing, body posture and facial expression) increasing the difficulty in this context. 2) the additional context within the spoken language assists the model. Overall we find the BLEU-1 and 4 scores increase by 2.28 and 1.42 when translating from text. Our approach circumvents the need for expensive gloss annotation, paving the way for better communication with the deaf community.

\subsection{Qualitative Evaluation}
\begin{figure}[!ht]
    \centering
    \caption{A Translation example produced by our best model on the RWTH-PHOENIX-Weather-2014\textbf{T} dataset.}
    \includegraphics[width=0.48\textwidth]{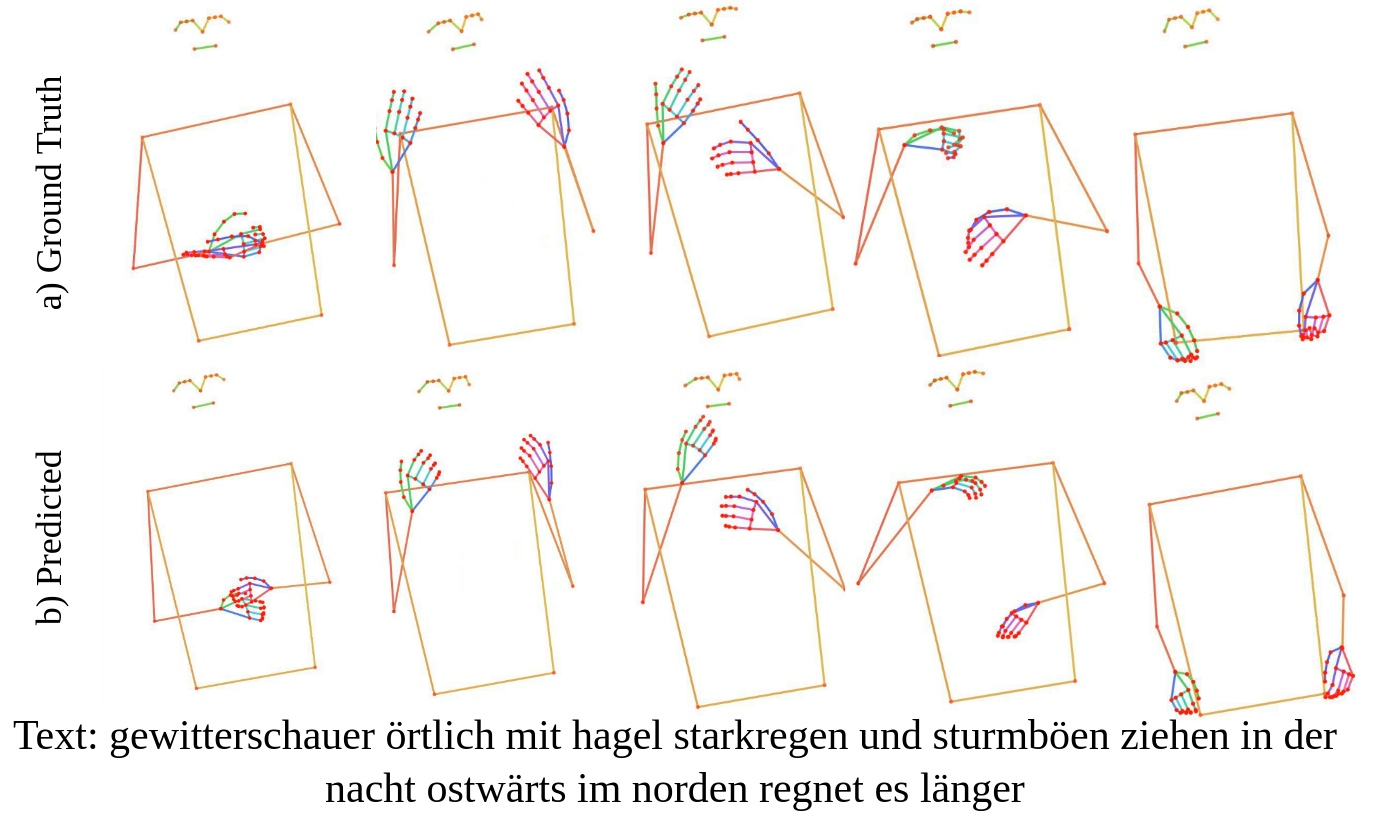}
    \label{fig:skeleton_translation_examples}
\end{figure}

\cref{fig:skeleton_translation_examples} shows a translation example from the \ac{ph14t} dataset. The figure shows the model is able to faithfully translate a given sentence. However, note that some details are missing in the hands caused by the quantization error from the codebook.

In addition, we share video outputs from our best models on the \ac{mdgs} and \ac{ph14t} datasets\footnote{\url{https://github.com/walsharry/VQ_SLP_Demos}}. To ensure a realistic evaluation we also share failure cases. Finally, we share PCA plots of the codebook's embedding space, showing the effect of our replacement strategy and contrastive loss. 

\vspace{-5mm}

\section{CONCLUSION}
\label{sec:conclusion}
In this work, we presented a novel approach to \ac{ttp} translation. Previously the task was treated as a pose regression problem, where the goal was to synthesize a pose sequence directly from text. As a result, the resulting poses suffered from regression to the mean. Here we propose performing a discrete sequence-to-sequence translation using a transformer. To accomplish this we create a discrete representation of sign language, in which the tokens can be combined to create continuous natural expressive signing. We explored the application of sign stitching to generate seamless, more natural sequences. Furthermore, we showed how linguistic annotation can be leveraged to improve our approach. In cases where linguistic annotation is absent, we demonstrated the feasibility of sharing codebooks across datasets.
We evaluate our approach on the \ac{ph14t} and \ac{mdgs} dataset, showing state-of-the-art back translation and \ac{dtwmje} scores.

\newpage





{\small
\bibliographystyle{ieee}
\bibliography{egbib}
}

\end{document}